\documentclass[10pt, conference, letterpaper]{IEEEtran}
\IEEEoverridecommandlockouts
\usepackage{cite}
\usepackage{amsthm}
\usepackage{amsmath,amssymb,amsfonts}
\usepackage[cmintegrals]{newtxmath}
\usepackage{algorithm}
\usepackage{algorithmicx}
\usepackage{enumerate}
\usepackage{algpseudocode}
\usepackage{caption}
\usepackage{bm}
\usepackage{subfig}
\usepackage{graphicx}
\usepackage{array}
\usepackage{textcomp}
\usepackage{xcolor}
\usepackage{url}
\usepackage{tabularx}
\usepackage{color}
\makeatletter

\usepackage{enumitem}
\setlist{nosep} 
\usepackage[font=small, skip=3pt]{caption}
\makeatother
\newlength{\whilewidth}
\algdef{SE}[parWHILE]{parWhile}{EndparWhile}[1]
  {\parbox[t]{\dimexpr\linewidth-\algmargin}{%
     \hangindent\whilewidth\strut\algorithmicwhile\ #1\ \algorithmicdo\strut}}{\algorithmicend\ \algorithmicwhile}%
\algnewcommand{\parState}[1]{\State%
  \parbox[t]{\dimexpr\linewidth-\algmargin}{\strut #1\strut}}

\newtheorem{definition}{Definition}
\newtheorem{theorem}{Theorem}
\newtheorem{lemma}{Lemma}

\newtheorem{Assumption}{Assumption}

\begin{document}
\title{The Impact Analysis of Delays in Asynchronous Federated Learning with Data Heterogeneity for Edge Intelligence}    

 \author{Ziruo Hao,
Zhenhua~Cui, 
Tao~Yang,~\IEEEmembership{Member,~IEEE,}
\\
Xiaofeng~Wu,
Hui~Feng,~\IEEEmembership{Member,~IEEE,}
and~Bo~Hu,~\IEEEmembership{Member,~IEEE}      

\thanks{Z. Hao, Z. Cui, T. Yang (Corresponding author), X. Wu, H. Feng and B. Hu are with the Department of
Electronic Engineering, School of Information Science and Technology,
Fudan University, Shanghai 200438, China (e-mail: zrhao22@m.fudan.edu.cn, czh773742344@sina.com, \{taoyang, xiaofeng\_wu, hfeng, bohu\}@fudan.edu.cn). H. Feng and B. Hu are also with the the Shanghai Institute of Intelligent Electronics and Systems, Shanghai 200433, China.
}
}
\maketitle

\begin{abstract}
Federated learning (FL) has provided a new methodology for coordinating a group of clients to train a machine learning model collaboratively, bringing an efficient paradigm in edge intelligence. Despite its promise, FL faces several critical challenges in practical applications involving edge devices, such as data heterogeneity and delays stemming from communication and computation constraints. This paper examines the impact of unknown causes of delay on training performance in an Asynchronous Federated Learning (AFL) system with data heterogeneity. Initially, an asynchronous error definition is proposed, based on which the solely adverse impact of data heterogeneity is theoretically analyzed within the traditional Synchronous Federated Learning (SFL) framework. Furthermore, Asynchronous Updates with Delayed Gradients (AUDG), a conventional AFL scheme, is discussed. Investigation into AUDG reveals that the negative influence of data heterogeneity is correlated with delays, while a shorter average delay from a specific client does not consistently enhance training performance. In order to compensate for the scenarios where AUDG are not adapted, Pseudo-synchronous Updates by Reusing Delayed Gradients (PSURDG) is proposed, and its theoretical convergence is analyzed. In both AUDG and PSURDG, only a random set of clients successfully transmits their updated results to the central server in each iteration. The critical difference between them lies in whether the delayed information is reused. Finally, both schemes are validated and compared through theoretical analysis and simulations, demonstrating more intuitively that discarding outdated information due to time delays is not always the best approach.
\end{abstract}

\begin{IEEEkeywords}
Asynchronous Federated Learning, data heterogeneity, edge intelligence, delayed information reusing 
\end{IEEEkeywords}

\section{Introduction}
Nowadays, edge devices are continuously generating and storing massive amounts of data by interacting with the natural world \cite{khan2021federated}. This data is the foundational fuel for Artificial Intelligence (AI) applications\cite{yang2021edge,hilmkil2021scaling}. Given that the data is collected and saved separately among the edge devices, how to coordinate these devices to train the same targeted machine learning model is the key to applying AI in edge intelligence scenarios. FL, as an emerging paradigm in distributed learning, addresses this challenge with high efficiency, overcoming the limitation associated with the inaccessibility of locally preserved data in each edge device \cite{lim2020federated}. FL has been applied in variant edge scenarios, which has proved promising with many successful implementations, such as the virtual keyboard project developed by Google AI \cite{hard2018federated} and autonomous driving \cite{you2023federated}. Moreover, FL brings the AI closer to where the data is generated, which diminishes the significance of a center that traditionally bears the substantial computational burden in a centralized training manner \cite{lim2020federated}. Meanwhile the effectiveness of the traditional SFL system is validated through theoretical analysis and simulations \cite{zhang2021survey}

However, the effectiveness of FL dramatically relies on the synchronous scheduling of the clients and the assumption of independent and identically distributed statistic characteristics of locally preserved data. Unfortunately, meeting these two conditions proves challenging in the practical implementation of FL within edge intelligence networks, as highlighted in \cite{yu2021toward}. In the edge network, the local data collected and preserved by each client are inevitably Non-Independent and Identically Distributed (Non-IID) due to different deployment locations and various data collection conditions \cite{hsieh2020non}. The adverse impact of data heterogeneity has long been recognized as a persistent challenge within FL. Apart from the data heterogeneity issue, the absence of each client also brings significant performance degradation, even making the global parameters non-convergent \cite{zhao2018federated}. The cost of keeping the clients synchronous is usually unacceptable, showing low training efficiency since the center has to wait for the slowest client, referred to as the straggler problem \cite{chen2018lag}. Besides, ensuring client synchronization is sometimes impractical in demanding environments with limited resources, such as the limited number of available communication channels \cite{amiri2021convergence}. Delays exacerbate the negative impact of data heterogeneity; hence, theoretical and experimental analyses of the extent of this negative effect are vital to improving the training performance of FL deployment in edge networks.

In the AFL, the delay varies during the learning process, referring to the number of absent training iterations for the corresponding client \cite{hu2021device}. In detail, the delay increments by one if the corresponding client fails to transmit the local results in the current iteration, while it resets to zero upon the successful transmission to the center \cite{cui2023data}.  There are various AFL schemes designed to adapt to different scenarios. They share two common points: first, not all clients participate in every training iteration due to transmission failures and limited computational capacity; and second, the server only uses the most recent local update results\cite{xu2023asynchronous}. In detail, if global parameters are calculated utilizing the local parameters updated from the global parameters in the last iteration, no outdated information is involved in the training process\cite{nguyen2022bufferAFL}.

Another variant of the AFL scheme involves the center sending global parameters to clients that have successfully transmitted their local results in the current iteration, without the need for selecting or waiting for specific clients. This approach is straightforward to implement and maximizes the participation of all clients to accelerate the training process, as demonstrated in \cite{hu2023scheduling}. Consequently, this paper focuses on and explores this particular scheme. However, a significant challenge arises in that the global parameters are updated using delayed local parameters from a varying subset of clients, which inevitably introduces outdated information into the aggregation process, degrading the training performance. Therefore, a critical aspect of this approach is how to handle these incomplete and delayed local parameters—referred to as the aggregation rule—to maintain training effectiveness. Interestingly, our subsequent analysis reveals that this delayed information does not always negatively impact training performance, offering a novel perspective on designing effective aggregation rules.

In this paper, we refer to the above asynchronous algorithm updates with delayed gradients, not actively using outdated information, as AUDG. Different from AUDG, this paper proposes a new class of pseudo-synchronous algorithms that update by reusing delayed gradients called PSURDG. Generally, they can be distinguished depending on whether the local parameters used in the aggregation step contain outdated information. The performance degradation due to data heterogeneity already exists in the SFL. Moreover, introducing delays in the AFL further exacerbates this adverse impact. This paper conducts a comprehensive analysis of the data heterogeneity and delays jointly to reveal their isolated and correlated impacts on the training performance of two AFL schemes with different aggregation rules, AUDG and PSURDG. The AUDG is construed through a conventional training paradigm, characterized by solely utilizing received information within each iteration. Conversely, PSURDG proposes a novel approach to dealing with striking data heterogeneity scenarios. The main contributions of this paper are concluded as follows:

\begin{itemize}
\item  We proposed an asynchronous error definition, which can be used to measure the negative influence of clients' deficiency caused by objective constraints. Based on the above, the convergence of SFL with Non-IID local data sets is theoretically proved. The negative influence of data heterogeneity is shown to slow down the convergence.

\item The convergence analysis of AUDG is presented, where the center only uses the received information to update the global parameters. Additionally, the associated theoretical analysis indicates that a smaller average delay of one specific client does not always lead to an overall better training performance, which is rarely discussed in previous works.    

\item A novel scheme, PSURDG, is proposed with convergence analysis to eliminate the influence of communication limitation on aggregation by increasing storage space. Furthermore, a comparative study with AUDG shows that the efficacy of reusing delayed information prevails in specific scenarios characterized by more minor delays and more extensive data heterogeneity. 
\end{itemize}

\begin{figure*}[!t]
\centering
\includegraphics[width=6in]{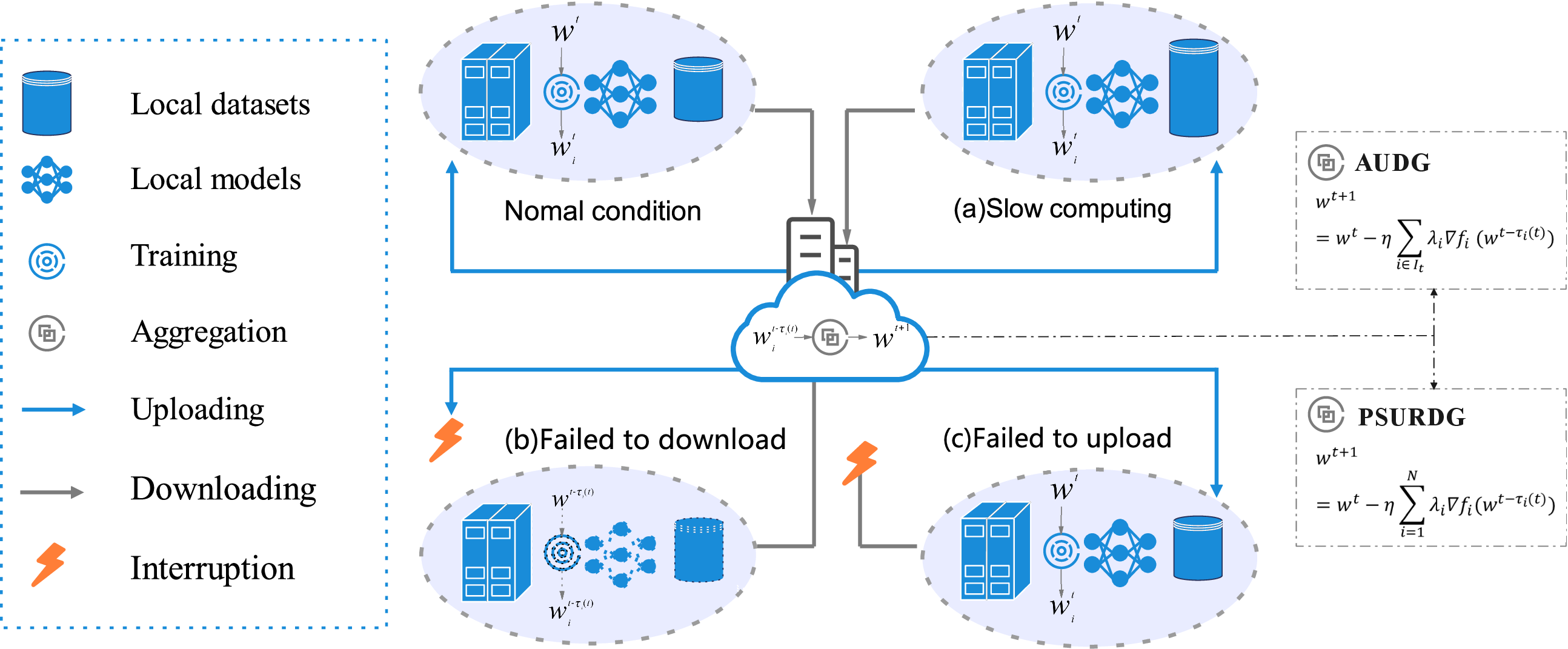}
\captionsetup{font={small}}
\caption{Delays in an asynchronous federated system can be attributed to three primary factors: (a)Slow computing speed (b)The failure to download global parameters (c)The failure to upload local parameters.}
\label{AFLscheme}
\end{figure*} 

\section{Related Work}
A common solution of asynchronism involves selecting a subset of clients at the start of each training iteration and sending them the latest global parameters. Such a solution prevents outdated information from influencing the global parameter updates, as described in \cite{cui2023data} and \cite{zhu2022online}. Once these clients have transmitted their updates, the central server aggregates the received results to update the global parameters. However, the success of this approach largely depends on selecting the right clients in each iteration, which requires a detailed understanding of each client's conditions and environment \cite{zhang2021client}. This requirement can be challenging to meet. Additionally, the central server still has to wait for the slowest client, which leads to inefficiencies, particularly in AFL with a large number of clients.

To reduce the server's waiting time, some studies, such as \cite{wang2022accelerating}, \cite{zhang2023fedmds}, and \cite{wu2023hiflash}, propose hierarchical aggregation structures. These approaches introduce intermediate edge servers that perform local aggregations before the final global aggregation, reducing the burden on the central server. However, deploying such hierarchical schemes is inherently complex and poses additional challenges.

Client selection remains a widely used method to improve AFL performance. For instance, \cite{nishio2019client} selects clients based on their communication conditions, with clients experiencing poor connectivity excluded from the training. However, this approach overlooks the issue of data heterogeneity. In contrast, \cite{zhu2022client} defines the delay of each client as "staleness," using it as an indirect measure of training performance and setting staleness as an optimization target. This method, however, assumes that shorter delays always result in better training performance—an assumption that lacks theoretical validation. Moreover, the effectiveness of such client selection mechanisms relies on the central server's ability to continuously monitor all clients, which is difficult, especially in large-scale networks.

In addition to modifying the FL structure, several studies focus on changing the aggregation rules in AFL to avoid the need for the central server to actively select clients and wait for their responses. For example, \cite{you2022triple} proposes a new aggregation method using a temporal weight fading strategy, though its effectiveness is only demonstrated in simulations. Similarly, \cite{wang2022gradient} suggests a partial averaging mechanism to schedule the received updates, but this approach also lacks theoretical analysis. Although some works, like \cite{zhou2022towards}, consider both data heterogeneity and delays, the relationship between these two factors remains largely unexplored.

It is well established that delays and data heterogeneity significantly degrade training performance in FL, particularly in resource-constrained edge environments \cite{imteaj2021survey}. While performance degradation due to data heterogeneity has been extensively discussed in many studies, such as \cite{zhu2021federated}, most of these studies focus on non-delay SFL. Although some methods, like \cite{xie2019asynchronous}, manage stale gradients, they often fail to address the combined impact of data heterogeneity and delays.

\section{The System Model of Asynchronous Federated Learning Scheme}
\subsection{The System Model}
\begin{figure*}[!t]
\centering
\includegraphics[width=7in]{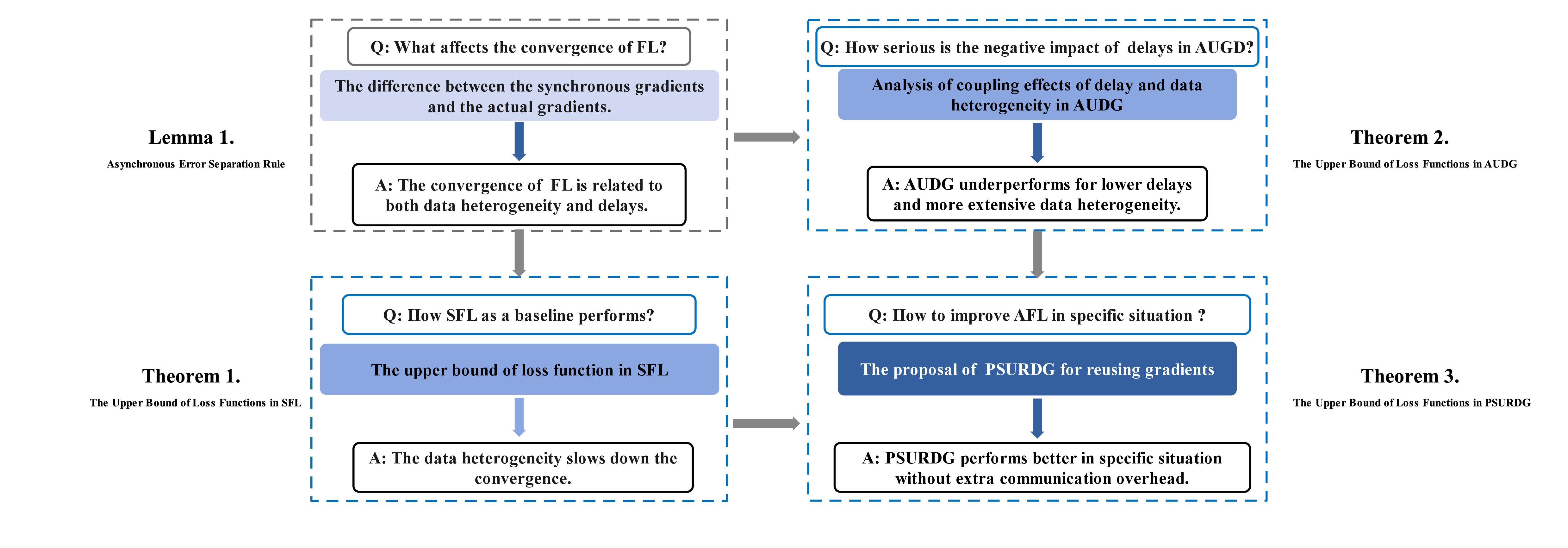}
\captionsetup{font={small}}
\caption{The theoretical analysis structure}
\label{analysisstructure}
\end{figure*} 
Apart from the center, there are $N$ clients formatting the FL network, denoted as $\mathcal{N}$. Given the diverse computational capabilities and stochastic communication channels inherent to each client, not all clients can transmit their updated local parameters synchronously, leading to asynchronous aggregation at the center. In this paper, we focus on the scenario where the center does not wait for the feedback of any clients, and it sends the latest global parameters to the client that just transmits its updated parameters successfully. Clients and servers train and aggregate at their own pace, keeping each other's latency known through upload and delivery timestamps. The whole workflow of the discussed AFL is shown in Fig. \ref{AFLscheme}.
Under this AFL scheme, the local parameters employed for global aggregation are updated based on the outdated global parameters with delays in relation to the current iteration. A random variable, denoted as $\tau_i(t)$, represents the delay associated with $client_i$.

\begin{equation}
\tau_i(t) =
\begin{cases}
  0 & \text{if } i \in \mathcal{I}_{t-1} \\
  \tau_i(t-1) + 1 & \text{if } i \notin \mathcal{I}_{t-1} \\
  t - \tau_i & \text{if } i \in \mathcal{I}_{t} \ (\text{Adjustment})
  \label{It}
\end{cases}
\end{equation}

$\tau_i$ is a timestamp to indicate which iteration $client_i$ is based on for successfully updating its gradient to the center in $t^{th}$ iteration. $\tau_i(t)$ is equal to $\tau_i(t-1) + 1$ or $0$ in most cases. Only in the case of downloading failure will $\tau_i(t)$ need to be adjusted with $\tau_i$. When $\tau_i(t) = 0$ for all the clients all the time, it represents the SFL scenarios. In practice, the iteration is a pre-defined time interval, and any clients that send their updated parameters in this interval are supposed to receive the same latest global parameters at the beginning of the next iteration. Furthermore, we define $\mathcal{I}_t$ as the clients that successfully send the updated parameters to the central server in $t^{th}$ iteration. Besides, we let $|\mathcal{I}_t|$ denote the number of this set, where $0\leq |\mathcal{I}_t| \leq N$. Following the above definitions, $\mathcal{I}_t$ is correlated with delays, but $|\mathcal{I}_t|$ is irrelevant with $\tau_i(t)$. It's worth noting that both $\tau_i(t)$ and $\tau_i$ are used to analyze the impact of latency on convergence. In real scenarios, clients and servers do not need $\tau_i(t)$ and $\tau_i$ for aggregation or training.


By following the standard definitions in FL, we let $\mathcal{D}_i$ represent the local data set from $client_i$, and $D_i=|\mathcal{D}_i|$ denotes the number of samples in this data set. Specifically, $w$ depicts the parameters and the loss function $f_i(w)$ from $client_i$ is defined as
\begin{equation}
f_i(w_i)={\frac{1}{D_i}}{\sum_{x\in{\mathcal{D}_i}}}\mathcal{L}_i(x;w_i), \label{local loss function}
\end{equation} 
where $\mathcal{L}_i(x;w_i)$ is the loss function value with data sample $x$ from $\mathcal{D}_i$ and parameters $w_i$. Among Non-IID data sets, the local optimal parameters for each local loss function $f_i$ are different and determined by the data characteristic associated with their respective local data sets, which is defined as
\begin{equation}
w_i^* = \text{argmin} \ f_i(w_i). \\    
\end{equation}
Additionally, $w_i^*$ and the theoretical optimal parameters $w^*$ may be different, and this difference, referred to as data dissimilarity between local and global data sets, significantly influences the training performance, which is discussed and analyzed by many existing works, such as \cite{wu2021fast} and \cite{chen2020asynchronous}. Under the AFL scheme, although $w^*$ cannot be obtained due to distributed data storage, it can be attained by optimizing the global loss function $f(w)$ in the centralized scheme. $f(w)$ is defined as the weighted average of $f_i(w)$, shown as follows 
\begin{equation}
    w^* = \text{argmin} \ f(w) ,
\end{equation}
\begin{align}
f(w) = \sum_{i=1}^N \lambda_i f_i(w), \; \text{where} \sum_{i=1}^N \lambda_i = 1. \label{global loss function}
\end{align}
$\lambda_i$ denotes the weight of $client_i$, and it signifies the importance of $client_i$'s local loss function, which can be interpreted as the normalized contribution of individual clients to the learning process. Our prior research has pointed out that $\lambda_i$ can be determined by both the data volume and statistic characteristics, which can be obtained through various approaches, such as the Shapley Value \cite{liu2022gtg}, if local data sets are Non-IID. 

In the domain of FL, the distance between global and local parameters is widely adopted as the metric to assess the heterogeneity between global and local data sets, such as \cite{yeganeh2020inverse}, and we characterize data heterogeneity by measuring the distance between local and global optimal parameters, denoted as ${\|w^*_i - w^*\|}^2$ in this paper. To ensure convergence, a assumption to bound the data heterogeneity among local data sets is presented below, and this assumption can be easily satisfied by applying pre-processing techniques to local data sets.

\begin{Assumption} (Bounded Data Heterogeneity):
\\
For each client, there exists a constant $\phi \geq 0$, such that
\begin{equation}
    \|w^*_i - w^*\|^2 \leq \phi^2
\end{equation}
\end{Assumption}

During local updating process, each client follows the following computation steps, where the learning rate $\eta$ is the same for all clients. 
\begin{equation}
    w_i^{t+1-\tau_i(t)} = w^{t-\tau_i(t)} - \eta \nabla f_i(w^{t-\tau_i(t)}),
\end{equation}
where $w^{t-\tau_i(t)}$ is the global parameters sent from the center at $(t-\tau_i(t))^{th}$ iteration, and $w_i^{t+1-\tau_i(t)}$ denotes the updated parameters transmitting successfully to the center with $\tau_i(t)$ delays in ${(t+1)}^{th}$ iteration.

In the aggregation process, the center receives incomplete and delayed gradients for global parameters updating in each iteration and uses them to obtain the latest global parameters shown in \eqref{global}, where the delayed gradient $\nabla f_i(w^{t-\tau_i(t)})$ can be obtained by simple subtraction between $w_i^{t+1-\tau_i(t)}$ and $w^{t-\tau_i(t)}$. 
\begin{equation}
    w^{t+1} = h (w_i^{{t+1-\tau_i(t)}},\;.\;.\;.\;,w_j^{{t+1-\tau_j(t)}}) \label{global}
\end{equation}
The mapping $h$, the aggregation rule, should be designed carefully, considering both delays and data heterogeneity. In this paper, we consider two AFL schemes with different aggregation rules, namely AUDG and PSURDG, whose client algorithms are delineated in Algorithm \ref{client algorithm}, and their aggregation rule are indicated by Algorithm \ref{AUDG algorithm} and Algorithm \ref{PSURDG algorithm}, respectively. AUDG is compatible with most current AFL models\cite{li2023asynchronous}, and PSURDG is proposed in this paper. 
\begin{algorithm}[!htbp]
  \captionsetup{font={small}}
  \caption{\emph{ClientUpdate}}
  \label{client algorithm}

  \begin{algorithmic}[1]
  \If{$t<T$}
  \For {all clients: $i={1,2,..,N}$}
  \If{Receive parameters $w^{t}$ and $t$ \textbf{and} standby}
  
  \State \textbf{Computing }the gradient $\nabla f_i(w^{\tau_i})$

  \State \textbf{Sends} $\nabla f_i(w^{\tau_i})$ to the center repeatedly
  \EndIf
  \EndFor
  \EndIf

  \end{algorithmic}

\end{algorithm}
\subsection{The Profile of Two Schemes and Notations}

The theoretical analysis structure is shown in Fig.\ref{analysisstructure}, and the convergence properties of all discussed cases are analyzed by deriving the upper bound of loss functions. At first, the convergence of SFL with Non-IID data sets is investigated to set the benchmark for the following two AFL schemes, namely AUDG and PSURDG. When the average delay equals zero across all the clients, the upper bound of AUDG and PSURDG are all equal to the SFL ones, thereby demonstrating the consistency of the theoretical analysis. As to the AFL discussed in this paper, AUDG can be seen as the normal way to deal with the absence of stochastic clients and delays in each iteration. One of the most notable features is the correlation between the negative influence of data heterogeneity and delays, and this correlation is detrimental to scenarios with large data heterogeneity.
To solve this problem, a novel AFL scheme, PSURDG, is proposed to decouple this correlation. The training performance of AUDG and PSURDG are compared via their upper bounds, thereby indicating their respective suitable application scenarios. To show the analysis clearly, Table.\ref{notations} summarizes several notations with their definitions.

\begin{table}[t]
\centering
\captionof{table}{Summary of Notations}
\label{notations}
\renewcommand\arraystretch{1.5}
\begin{tabular}{c|l}
 \hline
    \textbf{Notation}                                             & \textbf{Definition}\\ \hline
    $w^t$  
            & The global parameters at $t^{th}$ training iteration\\ \hline
    $w^t_i$                      
            & The local parameters of $clinet_i$ at $t^{th}$ training iteration   \\ \hline
    $w^*$
            & The global optimal parameters \\ \hline
    $w^*_i$
            & The local optimal parameters of $clinet_i$ \\ \hline
    $\eta$                             
            & Learning rate \\ \hline
    $\lambda_i$
            & The weight of $clinet_i$ \\ \hline

    $\tau_i(t)$
            & The delay of $clinet_i$ at $t^{th}$ training iteration\\ \hline
    $\tau_i$
            & The local time labels of $clinet_i$ at each iteration\\ \hline
    $\mathcal{I}_t$ & The set of successful transmission clients at $t^{th}$ iteration \\ \hline        
    $T$
            & The total number of communication rounds\\ \hline
    $\phi$                         
            & The parameter for bounded data heterogeneity assumption\\ \hline
    $L$                         
            & The parameter for smoothness assumption\\ \hline
    $\mu$
            & The parameter for convexity assumption \\ \hline
    $R$
            & The parameter for compactness assumption\\ \hline
    $G$
            & The parameter for bounded gradient assumption\\ \hline
    
\end{tabular}
\end{table}

\section{Asynchronous Updates with Delayed Gradients}
In this section, the AFL scheme AUDG is comprehensively analyzed, in contrast to PSURDG analyzed later, the delay gradient information of AUDG is used only once. The SFL is set as the benchmark, and its rigorous convergence proof with Non-IID data sets is provided, where the negative influence of data heterogeneity is theoretically discussed. The convergence analysis is also performed for the AUDG, revealing the performance degradation due to delays and data heterogeneity. We also note that the following analysis is based on gradient descent but can be seamlessly extended to stochastic gradient descent for more flexible implementation.

\begin{algorithm}
  \captionsetup{font={small}}
  \caption{\emph{AUDG ServerAggregation}}
  \label{AUDG algorithm}

  \begin{algorithmic}[1]
  \linespread{1}\selectfont
  \State \textbf{Initialization:} 
  \State Set $t=1$
  \State {The center sends the initialized model parameters $w^1$ and $t=1$ to all clients}
  \State \textbf{Center Side:}
  \For {$t < T+1$}
  \State \textbf{WAIT} for ClientUpdate

  \State \hspace{0.5cm} AUDG: $w^{t+1} = w^{t} - \eta \sum_{i \in \mathcal{I}_t} \lambda_i \nabla f_i(w^{t-\tau_i(t)})$ . 
  \State \textbf{Set $t = t+1$}
  \State \textbf{Transmits} the global parameters $w^{t}$ and $t$ to clients that successfully sends the parameters.
  \EndFor
  \State \textbf{Transmits} the global parameters $w^{t}$ to all clients repeatedly until all clients have received the final global parameters.

  \end{algorithmic}

\end{algorithm}
\subsection{The Definition of Asynchronous Error and The Convergence Analysis of Synchronous Updates}
Before the detailed analysis, some assumptions are introduced. Initially, the local loss function is assumed to be smooth and convex\cite{xiao2023time,you2023broadband}.  
\begin{Assumption}(Smoothness):
\begin{equation}
f_i(w_1)-f_i(w_2) \le <\nabla{f_i(w_2)},{w_1-w_2}>+\frac{L}{2}{\|w_1-w_2\|}^2. \label{smoothness i}
\end{equation}
\end{Assumption}

\begin{Assumption} (Convexity):
\begin{equation}
f_i(w_1) - f_i(w_2) \geq <\nabla{f_i(w_2)},w_1-w_2> + \frac{\mu}{2} {\|w_1-w_2\|}^2.
\end{equation}
\end{Assumption}

Various approaches exist to convert neural networks into equivalent convex optimization problems solved through gradient descent. These methodologies aim to make the target loss function of neural networks have approximate convex characteristics, and one of these techniques is employing over-parameterized neural networks, as discussed in \cite{fang2022convex}. Since the global loss function (\ref{global loss function}) is defined as the weighted average of local loss functions, it is easily inferred that the global loss function is also smooth and convex. 

Besides, a standard compactness assumption \cite{agarwal2011distributed} and the bounded gradient\cite{avdiukhin2021federated} are also needed, which is a common assumption in the machine learning field to analyze the convergence.
\begin{Assumption} (Compactness): \label{A3}
\begin{equation}
\|w^t - w^*\|^2 \leq R^2
\end{equation}
\end{Assumption}
\begin{Assumption} (Bounded gradient): \label{A4}
\\
Each loss function $f_i$ is differentiable and there exists a constant $G >0$ such that
\begin{equation}
    \|\nabla f_i(w)\|^2 \leq G^2.
\end{equation}
\end{Assumption}
As to AUDG, the aggregation rule and asynchronous error are defined as follows.

\begin{definition} 
In AUDG, the aggregation rule is defined as:
\begin{align}
w^{t+1} &= w^{t} + \sum_{i \in \mathcal{I}_t} \lambda_i ( w_i^{t+1-\tau_i(t)} - w^{t-\tau_i(t)}) \notag \\
&= w^{t} - \eta \sum_{i \in \mathcal{I}_t} \lambda_i \nabla f_i(w^{t-\tau_i(t)}), \label{AFL aggregation rules 1}
\end{align}
The asynchronous error at $t^{th}$ iteration, denoted by $e(t)$, is defined as the gradient difference between the synchronous case and the applied gradients $\sum_{i \in \mathcal{I}_t} \lambda_i \nabla f_i(w^{t-\tau_i(t)})$ under the proposed aggregation rule at $t^{th}$ iteration.
\begin{equation}
e(t) = \nabla f(w^t) - \sum_{i \in \mathcal{I}_t} \lambda_i \nabla f_i(w^{t-\tau_i(t)})  \label{asynchronous error 1}   
\end{equation} 
where $\nabla f(w^t)$ signifies the gradients for synchronous updating, while $\sum_{i \in \mathcal{I}_t} \lambda_i \nabla f_i(w^{t-\tau_i(t)})$ denotes the applied delayed gradients under the proposed aggregation rule.
\end{definition}
According to (\ref{AFL aggregation rules 1}), the update of global parameters, namely the aggregation step, can be interpreted as utilizing a set of delayed gradients. In order to evaluate the efficacy of the training process, the training performance is quantitatively assessed by deriving an upper bound for the loss function gap between the average of the updated parameters and $w^*$, which is also utilized in \cite{agarwal2011distributed} to analyze the convergence property in the field of learning algorithms. At first, the following lemma is presented to illustrate the impact of $e(t)$ on training performance. 
\begin{lemma} (Asynchronous Error Separation Rule) 
\\
Under the AFL scheme, the loss function $f$ has the following inequality after $T$ iterations 
\begin{align}
&\sum_{t=1}^{T} [f(w^{t+1}) - f(w^*)] \leq \frac{1}{2\eta}(\|w^1-w^*\|^2 - \|w^{T+1} - w^*\|^2) \notag \\
&+  \sum_{t=1}^{T} <e(t),w^{t+1} - w^*> +\frac{1}{2}(L - \frac{1}{\eta}) \sum_{t=1}^{T} \|w^{t+1} - w^{t}\|^2  \text{,} \label{lemma convergence 1}
\end{align}
where $w^{1}$ is the initialization parameters.
\begin{IEEEproof}
According to the convexity and smoothness assumption, we have the following inequality
\begin{align}
&f(w^t) - f(w^*) \leq <\nabla f(w^t),w^t - w^*> - \frac{\mu}{2} \|w^{*} - w^{t}\|^2 \notag \\
&\leq <\nabla f(w^t),w^{t+1} - w^*> + <\nabla f(w^t),w^t - w^{t+1}> \notag \\
&\leq <\nabla f(w^t),w^{t+1} - w^*> + f(w^t) - f(w^{t+1}) 
\notag \\
&+ \frac{L}{2} \|w^{t+1} - w^{t}\|^2.
\end{align}
By rearranging the above inequality, we have 
\begin{equation}
f(w^{t+1}) - f(w^*) \leq <\nabla f(w^{t}),w^{t+1} - w^*> + \frac{L}{2} \|w^{t+1} - w^{t}\|^2. \label{lemma step 1}
\end{equation}
By substituting $e(t) = \nabla f(w^t) - \sum_{i \in \mathcal{I}_t} \lambda_i \nabla f_i(w^{t-\tau_i(t)})$ into (\ref{lemma step 1}), the following inequality is obtained
\begin{align}
& f(w^{t+1}) - f(w^*) \leq <\sum_{i \in \mathcal{I}_t} \lambda_i \nabla f_i(w^{t-\tau_i(t)}),w^{t+1} - w^*> \notag \\
& + <e(t),w^{t+1} - w^*> + \frac{L}{2} \|w^{t+1} - w^{t}\|^2.
\end{align}
Recalling that $w^{t+1} = w^{t} - \eta \sum_{i \in \mathcal{I}_t} \lambda_i \nabla f_i(w^{t-\tau_i(t)})$, we have $\sum_{i \in \mathcal{I}_t} \lambda_i \nabla f_i(w^{t-\tau_i(t)}) = -\frac{1}{\eta}(w^{t+1} - w^{t})$, so that
\begin{align}
&f(w^{t+1}) - f(w^*)\notag \\ 
&\leq -\frac{1}{\eta} <w^{t+1}-w^{t},w^{t+1} - w^*> + <e(t),w^{t+1} - w^*> \notag \\
&+ \frac{L}{2} \|w^{t+1} - w^{t}\|^2 \notag \\ 
&\leq \frac{1}{2\eta}(\|w^t - w^*\|^2 - \|w^{t+1} - w^{*}\|^2) + <e(t),w^{t+1} - w^*> \notag \\
&+\frac{1}{2}(L - \frac{1}{\eta}) \|w^{t+1} - w^{t}\|^2. \label{lemma step 2} 
\end{align}
By summing (\ref{lemma step 2}) over $T$ iterations, this lemma is proved.
\end{IEEEproof}
\end{lemma}
\eqref{lemma convergence 1} shows the upper bound of the accumulated error between the loss function of global parameters and the theoretical optimal ones over $T$ iterations. It is evident that the asynchronous error $<e(t),w^{t+1} - w^*>$ in each iteration increases the upper bound \eqref{lemma convergence 1}, thereby indicating the performance degradation. The detrimental effect of asynchronism is fully concluded by $<e(t),w^{t+1} - w^*>$. By setting $<e(t),w^{t+1} - w^*> = 0$, the performance upper bound of SFL can be derived as follows, revealing its convergence characteristics, where the convergence is only impacted by data heterogeneity. The motivation of investigating the SFL at first is to set it as the benchmark for the AUDG and PSURDG.   

\begin{theorem} (The Upper Bound of Loss Functions in SFL with Data Heterogeneity)
\\
Under the SFL scheme, we have 
\begin{align}
f(\hat{w}(T)) - f(w^*) \leq \frac{R^2}{2\eta T} + \frac{2L}{\mu T^2}[ L R^2 + (\mu+L) \phi^2 ] \label{synchronous updates}
\end{align}
where $\hat{w}(T) = \frac{1}{T}\sum_{t=1}^T w^{t+1}$, representing the running average of output global parameters in the center.
\end{theorem}

Theorem 1 is proved in Appendix A. In \eqref{synchronous updates}, it is evident that the final global parameter converges to $w^*$ as the iteration $T$ increase. Compared with the IID scenarios, namely $w^* = w_i^*, \; \forall i \in \mathcal{N}$, the data heterogeneity leads to a higher upper bound indicated by $\phi^2$, showing inferior training performance. Though many existing works also discussed and proved the convergence of SFL with Non-IID data sets, we further show that the negative influence of data heterogeneity can be neutralized by increasing the number of iterations $T$. In other words, the data heterogeneity slows down the convergence speed, but the final global parameters still can be ensured to converge to $w^*$, which is consistent with the conclusion that the SFL is equivalent to the centralized training scheme with enough iteration times $T$. Nevertheless, this conclusion no longer holds when delays in the AFL scheme exist. The negative influence resulting from delays and the correlation between data heterogeneity and delays are further investigated as follows.

\subsection{Convergence Analysis of AUDG}
Under the aggregation rule (\ref{AFL aggregation rules 1}), the upper bound of the loss function between $\hat{w}(T)$ and $w^*$ in AUDG is given in the following part. Since $\tau_i(t)$ and $\mathcal{I}_t$ are both random variables, the expectation over total $T$ iterations is taken on both sides.

\begin{theorem} (The Upper Bound of Loss Functions in AUDG)
\\
Under the AFL scheme with the aggregation rule \eqref{AFL aggregation rules 1}, we have  
\begin{align}
&E[f(\hat{w}(T))] - f(w^*) \leq  \frac{R^2}{2\eta T} + \frac{2L}{\mu T^2 }[L R^2 + ( \mu +L) \phi^2] \notag \\
&+  \frac{L R^2}{2}  \sum_{i=1}^N \lambda_i E[\tau_i(t)] + (N - E[|\mathcal{I}_t|])(\frac{2L-\mu}{2} \phi^2 + \frac{3}{2} LR^2) \notag \\
& +\frac{\eta^2 G^2 (L-\mu)}{2} E[|\mathcal{I}_t|] \sum_{i=1}^N \lambda_i   E[\tau_i(t)] \notag \\
&+ \frac{\eta^2 G^2 LN}{2} \sum_{i=1}^N \lambda_i E [\frac{1}{3}\tau_i(t)^3 + \frac{3}{2}\tau_i(t)^2 + \frac{13}{6} \tau_i(t)] \label{upper bound delay gradients}
\end{align}

\begin{IEEEproof} 
According to the proposed asynchronous aggregation rules, the error at $t^{th}$ iteration is described as
\begin{equation}
    e(t) = \sum_{i=1}^N \lambda_i \nabla f_i(w^{t}) - \sum_{j \in \mathcal{I}_t} \lambda_j \nabla f_j(w^{t-\tau_j(t)}) .
\end{equation}
By equivalent transformation, we have
\begin{align}
&\sum_{t=1}^T <e(t),w^{t+1} - w^*> \notag \\
&= \sum_{i=1}^N \lambda_i  \sum_{t=1} ^T<\nabla f_i(w^t) - \nabla f_i(w^{t-\tau_i(t)}),w^{t+1} - w^*> \notag \\ 
&+  \sum_{i=1}^N \lambda_i \sum_{t=1}^T <\nabla f_i(w^{t-\tau_i(t)}),w^{t+1} - w^*>  \notag \\
&+  \sum_{t=1}^T <-\sum_{j \in \mathcal{I}_t} \lambda_j \nabla f_j(w^{t-\tau_j(t)}),w^{t+1} - w^*> 
\end{align}
According to the well-known inequality related to inner product and Bregman divergence, we have
\begin{align}
&<\nabla f(a) - \nabla f(b),c-d> \notag \\
&= D_f(d,a) - D_f(d,b) - D_f(c,a) + D_f(c,b) \label{Df}
\end{align}
where $D_f(a,b) = f(a) -f(b) -<\nabla f(b),a-b>$ denotes the Bregman divergence of loss function $f$. Consequently, $<\nabla f_i(w^t) - \nabla f_i(w^{t-\tau_i(t)}),w^{t+1} - w^*>$ can be decomposed as
\begin{align}
& <\nabla f_i(w^t) - \nabla f_i(w^{t-\tau_i(t)}),w^{t+1} - w^*> =  D_{f_i}(w^*,w^t)  \notag \\
&- D_{f_i}(w^*,w^{t-\tau_i(t)}) - D_{f_i}(w^{t+1},w^t) + D_{f_i}(w^{t+1},w^{t-\tau_i(t)}). \label{bregman}
\end{align}
Due to the convexity of $f_i$, $D_{f_i}$ is lower bounded
\begin{equation}
D_{f_i}(a,b) = f_i(a) - f_i(b) - <\nabla f_i (b),a-b> \geq \frac{\mu}{2} \|a-b\|^2.\label{D convexity}
\end{equation}
Due to the smoothness of $f_i$, $D_{f_i}$ is upper bounded
\begin{equation}
D_{f_i}(a,b) = f_i(a) - f_i(b) - <\nabla f_i (b),a-b> \leq \frac{L}{2} \|a-b\|^2. \label{D smoothness}
\end{equation}
According to \eqref{bregman}, $\sum_{t=1}^T <e(t),w^{t+1} - w^*> $ can be decomposed into three parts from A to C.
\begin{align}
&\sum_{t=1}^T <e(t),w^{t+1} - w^*>  \\
&= \sum_{i=1}^N \lambda_i  \sum_{t=1} ^T [D_{f_i}(w^*,w^t) - D_{f_i}(w^*,w^{t-\tau_i(t)})] \tag{A} \\
&- \sum_{i=1}^N \lambda_i  \sum_{t=1} ^T [D_{f_i}(w^{t+1},w^t) - D_{f_i}(w^{t+1},w^{t-\tau_i(t)})] \tag{B} \\ 
&+  \sum_{t=1}^T \sum_{j \notin \mathcal{I}_t} \lambda_j <\nabla f_j(w^{t-\tau_j(t)}),w^{t+1} - w^*> \tag{C} \\
&=A + B +C \notag
\end{align}
where $j \notin \mathcal{I}_t$ denotes the set of clients that fail to send parameters to the center at $t^{th}$ iteration. According to  Assumption \ref{A3}, part A is upper bounded as 
\begin{align}
A&= \sum_{i=1}^N \lambda_i \sum_{t=1}^T [D_{f_i}(w^*,w^t) - D_{f_i}(w^*,w^{t-\tau_i(t)})] \notag \\
&\leq  \frac{L R^2}{2} \sum_{i=1}^N \sum_{t=1}^T \lambda_i \tau_i(t) .\label{A upperbound}
\end{align} 

Then, we focus on the part B, and rearrange the term $D_{f_i}(w^{t+1},w^t) - D_{f_i}(w^{t+1},w^{t-\tau_i(t)})$ as
\begin{align}
&D_{f_i}(w^{t+1},w^t) - D_{f_i}(w^{t+1},w^{t-\tau_i(t)}) \notag \\
&= \sum_{s=t-\tau_i(t)}^{t-1}[D_{f_i}(w^{t+1},w^{s+1}) - D_{f_i}(w^{t+1},w^{s})] .\label{Dfi2}
\end{align}
According to the features of Bergman divergence \eqref{D convexity} and \eqref{D smoothness}, $D_{f_i}(w^{t+1},w^{s+1}) - D_{f_i}(w^{t+1},w^{s})$ is lower bounded, i.e.,  
\begin{align}
&D_{f_i}(w^{t+1},w^{s+1}) - D_{f_i}(w^{t+1},w^{s}) = f_i(w^s) - f_i(w^{s+1}) \notag \\ 
&- <\nabla f_i(w^{s+1}), w^{t+1} - w^{s+1}> + <\nabla f_i(w^{s}), w^{t+1} - w^{s}> \notag \\
& \geq <\nabla f_i(w^{s+1}), w^{s} - w^{s+1}> +\frac{\mu}{2} \|w^{s+1} - w^{s}\|^2\notag \\
&- <\nabla f_i(w^{s+1}), w^{t+1} - w^{s+1}> + <\nabla f_i(w^{s}), w^{t+1} - w^{s}> \notag \\
&\geq -\frac{L-\mu}{2} \|w^{s+1} - w^{s}\|^2 -\frac{L}{2}\|w^{t+1} - w^{s}\|^2 .
\end{align}
Jointly considering the convexity and smoothness assumption, we have $L \geq \mu$ and give the following inequality.
\begin{align}
&-[D_{f_i}(w^{t+1},w^t) - D_{f_i}(w^{t+1},w^{t-\tau_i(t)})] \notag \\
&\leq \sum_{s=t-\tau_i(t)}^{t-1} (\frac{L-\mu}{2} \|w^{s+1} - w^{s}\|^2 + \frac{L}{2} \|w^{t+1} - w^{s}\|^2) \label{b inter}
\end{align}

According to the bounded gradient assumption and inequality $\|\sum_{i=1}^N a_i\|^2 \leq N \sum_{i=1}^N \|a_i\|^2$, $\|w^{t+1} - w^t\|^2$ is bounded, i.e.,
\begin{align}
\|w^{t+1} - w^t\|^2 &= \|\eta \sum_{j \in \mathcal{I}_t} \lambda_j \nabla f_j(w^{t-\tau_j(t)}) \|^2 \notag \\ 
&\leq |\mathcal{I}_t| \eta^2 G^2. \label{w^{t+1}-w^t asynchronous}
\end{align}
Due to the inequality $\|\sum_{i=1}^N a_i\|^2 \leq N \sum_{i=1}^N \|a_i\|^2$ ,$\|w^{t+1} - w^{t-\tau_i(t)}\|^2$ is also bounded
\begin{align}
\|w^{t} - w^{t-\tau_i(t)}\|^2 &\leq \tau_i(t)\sum_{s=1}^{\tau_i(t)} \|w^{t+1-s} - w^{t-s}\|^2 \notag \\
&\leq (\tau_i(t))\eta^2 G^2 \sum_{s=1}^{\tau_i(t)}|\mathcal{I}_{t-s}|  \notag \\
&\leq (\tau_i(t))^2 N \eta^2 G^2 \label{ws}.
\end{align}

By applying (\ref{ws}) in \eqref{b inter}, we have
\begin{align}
&-[D_{f_i}(w^{t+1},w^t) - D_{f_i}(w^{t+1},w^{t-\tau_i(t)})] \notag \\
&\leq \frac{L-\mu}{2} \tau_i(t)|\mathcal{I}_t| \eta^2 G^2  + \frac{L}{2} N \sum_{s=t-\tau_i(t)}^{t-1}  (t+1-s)^2 \notag \\
& \leq \frac{\eta^2 G^2 }{2} [|\mathcal{I}_t| (L-\mu)\tau_i(t) + LN  (\frac{1}{6} (2\tau_i(t)+3)(\tau_i(t)+2) \notag \\
&(\tau_i(t)+1) -1)].
\end{align} 
By summing the above upper bound from each client, the part B has the following inequality. 
\begin{align}
&B = - \sum_{i=1}^N \lambda_i  \sum_{t=1} ^T [D_{f_i}(w^{t+1},w^t) - D_{f_i}(w^{t+1},w^{t-\tau_i(t)})]  \notag \\
&\leq \frac{\eta^2 G^2 (L-\mu)}{2} \sum_{i=1}^N \lambda_i \sum_{t=1} ^T   [|\mathcal{I}_t|\tau_i(t)] \notag \\
&+ \frac{\eta^2 G^2 LN}{2} \sum_{i=1}^N \lambda_i \sum_{t=1} ^T  (\frac{1}{3}\tau_i(t)^3 + \frac{3}{2}\tau_i(t)^2 + \frac{13}{6} \tau_i(t))  \label{B upperbound}
\end{align}

As to part C, we deal with the term $<\nabla f_j(w^{t-\tau_j(t)}),w^{t+1} - w^*>$ at first, and its upper bound can be derived according to smoothness and convexity assumption. 
\begin{align}
&<\nabla f_j(w^{t-\tau_j(t)}),w^{t+1} - w^*>\notag \\
&\leq f_i(w^{t+1}) - f_i(w^*) - \frac{\mu}{2} \|w^{t+1} - w^{t-\tau_i(t)}\|^2  \notag \\
&+\frac{L}{2} \|w^{t-\tau_i(t)} - w^*\|^2 \label{C 1}
\end{align}

Then, we try to derive the upper bound of $f_i(w^{t+1})$. Let $w^*_i$ denotes the optimal parameters making $\nabla f_i(w^*) = 0$, and thus $f_i(w^t)$ is upper bounded as
\begin{align}
f_i(w^t) &\leq f_i(w^*_i) + <\nabla f_i(w^*_i),w^t - w^*_i> + \frac{L}{2}\|w^t - w^*_i\|^2 \notag \\
&\leq f_i(w^*_i) + L(\|w^t - w^*\|^2 + \|w^* - w^*_i\|^2) . \label{fi bound}
\end{align}
By substituting \eqref{fi bound} into \eqref{C 1}, we have
\begin{align}
&<\nabla f_j(w^{t-\tau_j(t)}),w^{t+1} - w^*> \notag \\
& \leq \frac{2L-\mu}{2} \|w^* - w^*_j\|^2 + \frac{3}{2} LR^2 \label{key C}
\end{align}
Considering the bounded data heterogeneity assumption, the upper bound of part C is described as 
\begin{align}
C &\leq \sum_{t=1}^T \sum_{j \notin \mathcal{I}_t} \lambda_j (\frac{2L-\mu}{2} \|w^* - w^*_j\|^2 + \frac{3}{2} LR^2) \notag \\
&\leq \sum_{t=1}^T \sum_{j \notin \mathcal{I}_t} \lambda_j (\frac{2L-\mu}{2} \phi^2 + \frac{3}{2} LR^2) \notag \\
&\leq \sum_{t=1}^T (N - |\mathcal{I}_t|)(\frac{2L-\mu}{2} \phi^2 + \frac{3}{2} LR^2)
\label{C upperbound}
\end{align}

Finally, we add (\ref{A upperbound}), (\ref{B upperbound}), and (\ref{C upperbound}) together to give the upper bound of $\sum_{t=1}^T <e(t),w^{t+1} - w^*>$, i.e.,
\begin{align}
&\sum_{t=1}^T <e(t),w^{t+1} - w^*> \notag \\
&\leq \frac{L R^2}{2}  \sum_{t=1}^T \sum_{i=1}^N \lambda_i \tau_i(t) + \sum_{t=1}^T (N - |\mathcal{I}_t|)(\frac{2L-\mu}{2} \phi^2 + \frac{3}{2} LR^2) \notag \\
& + \frac{\eta^2 G^2 (L-\mu)}{2} \sum_{i=1}^N \lambda_i \sum_{t=1} ^T   [|\mathcal{I}_t|\tau_i(t)] \notag \\
&+ \frac{\eta^2 G^2 LN}{2} \sum_{i=1}^N \lambda_i \sum_{t=1} ^T  (\frac{1}{3}\tau_i(t)^3 + \frac{3}{2}\tau_i(t)^2 + \frac{13}{6} \tau_i(t)) \label{et1}
\end{align}
By adding \eqref{et1} with the upper bound of synchronous case and taking the expectation of both sides, we have 
\begin{align}
&E[f(\hat{w}(T))] - f(w^*) \leq  \frac{R^2}{2\eta T} + \frac{2L}{\mu T^2 }[L R^2 + ( \mu +L) \phi^2] \notag \\
&+  \frac{L R^2}{2}  \sum_{i=1}^N \lambda_i E[\tau_i(t)] + (N - E[|\mathcal{I}_t|])(\frac{2L-\mu}{2} \phi^2 + \frac{3}{2} LR^2) \notag \\
& +\frac{\eta^2 G^2 (L-\mu)}{2} E[|\mathcal{I}_t|] \sum_{i=1}^N \lambda_i   E[\tau_i(t)] \notag \\
&+ \frac{\eta^2 G^2 LN}{2} \sum_{i=1}^N \lambda_i E [\frac{1}{3}\tau_i(t)^3 + \frac{3}{2}\tau_i(t)^2 + \frac{13}{6} \tau_i(t)].
\end{align}
At this stage, the proof is completed. We also note that the expectation over $|\mathcal{I}_t|$ and $E[{\tau_i(t)}]$ can be decomposed, since $|\mathcal{I}_t|$ is determined by the $\tau_i(t+1)$ by definitions, which means $|\mathcal{I}_t|$ is independent with the $\tau_i(t)$.  
\end{IEEEproof}
\end{theorem}

When there are no delays throughout the learning process, the upper bound \eqref{upper bound delay gradients} is equal to the upper bound of SFL in \eqref{synchronous updates}. The introduction of delays hinders the above upper bound from reaching zero, suggesting that the parameters may not converge to the optimal ones.In \eqref{upper bound delay gradients}, some terms associated with delays remain invariant as $T$ increases, which implies that increasing $T$ does not mitigate the adverse impacts of delays. Generally, these delay-related terms underscore the negative effects from two aspects: delay-related data heterogeneity and pure delays. 
\vspace{3mm}
\newpage                                                                                                                                                                                                              
\textbf{The Negative Influence Analysis of Delay Correlated Data Heterogeneity:}

In AUDG, two notable imperfections relative to SFL are discernible: firstly, not all clients engage in updating the global parameters during each iteration, leading to different participating times in aggregation; secondly, the gradients utilized for global aggregation exhibit delays. 

In \eqref{upper bound delay gradients}, the term $(N - E[|\mathcal{I}_t|])\frac{2L-\mu}{2}\phi^2$ highlights the correlation between the data heterogeneity and the presence of delays. To further investigate the cause of the correlation between delays and data heterogeneity in AUDG, we first investigate the relationship between $E[|\mathcal{I}_t|]$ and $E[\tau_i(t)]$.
Generally, $E[|\mathcal{I}_t|]$ represents the degree of asynchrony. Clearly, although $\tau_i(t)$ and $|\mathcal{I}_t|$ are independent, $\tau_i(t)$ and $|\mathcal{I}_{t-1}|$ are related and hence the link between $E[\tau_i(t)]$ and $E(|\mathcal{I}_t|)$ exists. The relationship between $E[|\mathcal{I}_t|]$ and $E[\tau_i(t)]$ of any clients can be summarized as follows: 
\begin{align}
E[\tau_i(t)] \rightarrow{0}\;, E(|\mathcal{I}_t|) \rightarrow{E(|\mathcal{I}_t|) + 1} \label{It1} \\
E[\tau_i(t)] \rightarrow{T}\;, E(|\mathcal{I}_t|) \rightarrow{E(|\mathcal{I}_t|) - 1} \label{It2}
\end{align} 

 Delays from any clients exacerbate the adverse effects of data heterogeneity. This is evident as the term $(N - E[|\mathcal{I}_t|])\frac{2L-\mu}{2}\phi^2$ increases with a corresponding rise in $E[\tau_i(t)]$, which tends to decrease $E(|\mathcal{I}_t|)$. When $E(|\mathcal{I}_t|)$ remains constant, the increase in $\phi$ due to data heterogeneity further affects the convergence. In other words, the absence of some clients, indicated by $E[|\mathcal{I}_t|]$, aggravates the negative influence of data heterogeneity, which makes the global parameters biased towards the local optimal ones from clients that excessively participate in the training, thereby degrading the training performance.
\vspace{3mm}


\textbf{The Negative Influence Analysis only from Delays:}

Apart from the performance degradation due to delay correlated data heterogeneity, we further try to illustrate the negative influence only from delays and show how the delays from individual client solely affect the training performance. At first, we define Performance Degradation only due to Delays ($PDD$), constituted by the terms only related to delays in \eqref{upper bound delay gradients}. The term $PDD$ has two meanings: a) $PDD$ represents the upper bound of AUDG when $\phi = 0$ and $T \to \infty$; b) $PDD$ indicates the upper bound of the difference between AUDG and SFL for the same number of iterations.
\begin{align}
&PDD = \frac{L R^2}{2}  \sum_{i=1}^N \lambda_i E[\tau_i(t)] + \frac{3}{2}L R^2(N- E[|\mathcal{I}_t|])  \notag \\
&+\frac{\eta^2 G^2 LN}{2} \sum_{i=1}^N \lambda_i E [\frac{1}{3}\tau_i(t)^3 + \frac{3}{2}\tau_i(t)^2 + \frac{13}{6} \tau_i(t)] \notag \\
&+ \frac{\eta^2 G^2 (L-\mu)}{2} E[|\mathcal{I}_t|] \sum_{i=1}^N \lambda_i   E[\tau_i(t)]. \label{PDD1}
\end{align}
In the proposed scenario, each client is assumed to be allocated with an independent communication channel, and thus $E[\tau_i(t)]$ from each client is independent. As to the first three terms in $PDD$, a larger $E[\tau_i(t)]$ from any client would make them larger. However, clients are still mutually affected in training performance, as evidenced by $E[|\mathcal{I}_t|]$ determined by all clients. In particular, $E[|\mathcal{I}_t|]$ serves as a common multiplier for each client in the fourth term of (\ref{PDD1}), and the increase of $E[\tau_i(t)]$ of any clients may decrease the value of $\frac{\eta^2 G^2 (L-\mu)}{2} E[|\mathcal{I}_t|] \sum_{i=1}^N \lambda_i  E[\tau_i(t)]$, showing the correlation among all clients in the training process. This phenomenon can be explained as follows: For the client with an overly high average delay, its overly delayed gradient is harmful to the training performance. If its average delay cannot be decreased to a minimal value, increasing its average delay to reduce its participation times in the training process is better since its delay decreases to zero only when it sends the parameters, shown in \eqref{It}.  In other words, an increase in the average delay of one specific client, even forbidding a specific client from participating, may improve the training performance for some time, which is against the conclusion of many existing works, such as \cite{zhu2022client}.

Therefore, it is challenging to isolate the direct impact of individual client's average delays on training performance, since a decrease in the average delay of a particular client does not guarantee a consistently monotonic increase in training performance. Generally, it is more likely to find that a smaller average delay of one specific client always leads to a better training performance when the degree of data heterogeneity is large, which is verified in our simulations.

\section{Pseudo-synchronous Updates by Reusing Delayed Gradients} 
As to the scenario with large data heterogeneity, AUDG is intuitively unfeasible due to the correlation between the negative influence of data heterogeneity and delays. Therefore, a novel AFL scheme PSURDG is proposed, where the previously applied gradients are reused in each iteration. The corresponding convergence analysis is also conducted by deriving the upper bound of loss function value between $\hat{w}(T)$ and $w^{*}$. Comparing to AUDG, the PSURDG demonstrates a superior training performance in some specific scenarios.

\subsection{Convergence Analysis of Pseudo-synchronous Updates by Reusing Delayed Gradients}
In AUDG, we theoretically prove that the delay aggravates the negative effect of data heterogeneity on learning performance, which can be attributed to the different participation times of each client in the learning process. To mitigate this delay-related negative influence, especially for scenarios with large data heterogeneity, we heuristically let the center aggregate the gradients from all clients by reusing the delayed gradients information. To show the rationality and efficiency of this new scheme, the upper bound of loss function value between $\hat{w}(T)$ and $w^{*}$ is derived. Similarly to the previous section, we give the definition of the aggregation rule and corresponding asynchronous error at first. 

\begin{algorithm}
  \captionsetup{font={small}}
  \caption{\emph{PSURDG ServerAggregation}}
  \label{PSURDG algorithm}

  \begin{algorithmic}[1]
  \linespread{1}\selectfont
  \State \textbf{Initialization:} 
  \State Set $t=1$
  \State {The center sends the initialized model parameters $w^1$ and $t=1$ to all clients}
  \State \textbf{Center Side:}
  \For {$t < T+1$}
  \State \textbf{WAIT} for ClientUpdate

  \State \hspace{0.5cm} PSURDG: $w^{t+1} = w^{t} - \eta \sum_{i=1}^N \lambda_i \nabla f_i (w^{t-\tau_i(t)}) $. 
  \State \textbf{Set $t = t+1$}
  \State \textbf{Transmits} the global parameters $w^{t}$ and $t$ to clients that successfully sends the parameters.
  \EndFor
  \State \textbf{Transmits} the global parameters $w^{t}$ to all clients repeatedly until all clients have received the final global parameters.

  \end{algorithmic}

\end{algorithm}

\begin{definition}
In PSURDG, the aggregation rule is defined as:
\begin{align}
w^{t+1} = w^{t} - \eta \sum_{i=1}^N \lambda_i \nabla f_i(w^{t-\tau_i(t)}) , \label{AFL aggregation rules 2}
\end{align}
In this case, the asynchronous error at $t^{th}$ iteration, denoted by $e'(t)$, is defined as follows:
\begin{equation}
    e^{\prime}(t) = \nabla f(w^{t}) - \sum_{i=1}^N \lambda_i\nabla f_i(w^{t-\tau_i(t)}) \label{asyncrhonous error 2}
\end{equation}
where $ \nabla f(w^{t})$ denotes the gradient for updating in synchronous case. $\sum_{i=1}^N \lambda_i \nabla f_i(w^{t-\tau_i(t)})$ is constituted by two parts: the received delayed gradients and the previously applied gradients in the center.
\end{definition}
Referring to \eqref{AFL aggregation rules 2}, the center reuses the received gradients of the client that failed to transmit in the current iteration. Thus, the number of aggregation clients in \eqref{AFL aggregation rules 2} becomes $N$ compared to \eqref{AFL aggregation rules 1}.

We also note that Lemma 1 in \eqref{lemma convergence 1} is also satisfied under the aggregation rule (\ref{AFL aggregation rules 2}) of PSURDG, which can be easily proved by following the same proof steps in Lemma 1. Based on the above definitions, the upper bound under \eqref{AFL aggregation rules 2} is derived, and the expectation over total $T$ iterations is taken on both sides.
\begin{theorem} (Pseudo-synchronous Updates)
\\
Under the proposed Pseudo-synchronous learning scheme by reusing the delayed gradients, we have  
\begin{align}
&E[f(\hat{w}(T))] - f(w^*) \leq \frac{R^2}{2\eta T} + \frac{2L}{\mu T^2 }[L R^2 + ( \mu +L) \phi^2] \notag \\
& + \frac{N \eta^2 G^2(L-\mu)}{2} \sum_{i=1}^N \lambda_i E[\tau_i(t) + \frac{L}{L-\mu} (\frac{1}{3} \tau_i(t)^3 + \frac{3}{2}  \tau_i(t)^2 \notag \\
&+ \frac{13}{6} \tau_i(t))]  +  \frac{L R^2}{2}   \sum_{i=1}^N \lambda_i E[\tau_i(t)]\label{upper bound reusing gradients}
\end{align}
\begin{IEEEproof} 
According to the pseudo-synchronous aggregation rules, the error at $t^{th}$ iteration is described as follows.
\begin{equation}
    e^{\prime}(t) = \sum_{i=1}^N \lambda_i \nabla f_i(w^{t}) - \sum_{i=1}^N \lambda_j \nabla f_i(w^{t-\tau_i(t)})
\end{equation}
By equivalent transformation, we have
\begin{align}
&\sum_{t=1}^T <e^{\prime}(t),w^{t+1} - w^*> \notag \\
&= \sum_{i=1}^N \lambda_i  \sum_{t=1} ^T<\nabla f_i(w^t) - \nabla f_i(w^{t-\tau_i(t)}),w^{t+1} - w^*> \label{et2}
\end{align}
At first, \eqref{et2} is decomposed into two parts: $A^{\prime}$ and $B^{\prime}$.
\begin{align}
&\sum_{t=1}^T <e^{\prime}(t),w^{t+1} - w^*>  \\
&= \sum_{i=1}^N \lambda_i  \sum_{t=1} ^T [D_{f_i}(w^*,w^t) - D_{f_i}(w^*,w^{t-\tau_i(t)})] \tag{$A^{\prime}$} \\
&- \sum_{i=1}^N \lambda_i  \sum_{t=1} ^T [D_{f_i}(w^{t+1},w^t) - D_{f_i}(w^{t+1},w^{t-\tau_i(t)})] \tag{$B^{\prime}$} \\
&= A^{\prime} + B^{\prime} \notag
\end{align}

As to $A^{\prime}$, its upper bound can be derived by following the same steps of part A in the proof of Theorem 2. 
\begin{align}
A^{\prime}&= \sum_{i=1}^N \lambda_i \sum_{t=1}^T [D_{f_i}(w^*,w^t) - D_{f_i}(w^*,w^{t-\tau_i(t)})] \notag \\
&\leq  \frac{L R^2}{2} \sum_{i=1}^N \sum_{t=1}^T \lambda_i \tau_i(t) \label{A upperbound 2}
\end{align}

 As to part $B^{\prime}$, the proof idea is similar to the part B in \eqref{B upperbound}, but the upper bounds of $\|w^{t+1} - w^t\|^2$ and $\|w^{t+1} - w^{t-\tau_i(t)}\|^2$ are different. According to $\|\sum_{i=1}^N a_i\|^2 \leq N \sum_{i=1}^N \|a_i\|^2$, $\|w^{t+1} - w^t\|^2$ is bounded as 
\begin{align}
&\|w^{t+1} - w^t\|^2 = \|\eta \sum_{i=1}^N \lambda_i \nabla f_i(w^{t-\tau_i(t)}) \|^2 \notag \\ 
&\leq N \eta^2 \sum_{i=1}^{N} \|\lambda_i \nabla f_i(w^{t-\tau_i(t)})\|^2 \leq N \eta^2 G^2\label{w^{t+1}-w^t asynchronous2}.
\end{align}
Similarly, we have
\begin{align}
\|w^{t} - w^{t-\tau_i(t)}\|^2 &\leq \tau_i(t)\sum_{s=1}^{\tau_i(t)} \|w^{t+1-s} - w^{t-s}\|^2] \notag \\
&\leq (\tau_i(t))^2 N \eta^2 G^2
\label{ws2}.
\end{align}
 Then, the upper bound of $B^{\prime}$ can be described by following the similar steps in the proof of the upper-bound of AUDG.
\begin{align}
&B^{\prime} \leq \frac{N \eta^2 G^2 (L-\mu)}{2} \sum_{i=1}^N \lambda_i \sum_{t=1} ^T  [\tau_i(t) + \frac{L}{L-\mu}  (\frac{1}{6}(2\tau_i(t)+3) \notag \\
&(\tau_i(t)+2) (\tau_i(t)+1) -1)] \label{B upperbound 2}
\end{align}

Finally, we add (\ref{A upperbound 2}) and (\ref{B upperbound 2}) together to show the upper bound of $\sum_{t=1}^T <e^{\prime}(t),w^{t+1} - w^*>$.
\begin{align}
&\sum_{t=1}^T <e^{\prime}(t),w^{t+1} - w^*> \notag \\
&\leq  \frac{N \eta^2 G^2(L-\mu)}{2} \sum_{i=1}^N \lambda_i \sum_{t=1} ^T [\tau_i(t) + \frac{L}{L-\mu}  (\frac{1}{6} (2\tau_i(t)+3)\notag \\
&(\tau_i(t)+2) (\tau_i(t)+1) -1)]  + \frac{L R^2}{2}  \sum_{t=1}^T \sum_{i=1}^N \lambda_i \tau_i(t)\label{et3}
\end{align}
By adding \eqref{et3} with the upper bound of synchronous case and taking the expectation of both sides, we have
\begin{align}
&E[f(\hat{w}(T))] - f(w^*) \leq \frac{R^2}{2\eta T} + \frac{2L}{\mu T^2 }[L R^2 + ( \mu +L) \phi^2] \notag \\
& + \frac{N \eta^2 G^2(L-\mu)}{2} \sum_{i=1}^N \lambda_i E[\tau_i(t) + \frac{L}{L-\mu} (\frac{1}{3} \tau_i(t)^3 + \frac{3}{2}  \tau_i(t)^2 \notag \\
&+ \frac{13}{6} \tau_i(t))]  +  \frac{L R^2}{2}   \sum_{i=1}^N \lambda_i E[\tau_i(t)].
\end{align}
At this stage, the proof is completed.
\end{IEEEproof}
\end{theorem}

In \eqref{upper bound reusing gradients}, the introduction of delays also prevents the upper bound from converging to 0, indicating inferior training performance compared to the synchronous case. However, the strategy of reusing delayed gradients brings two significant new characteristics compared with not reusing strategy \eqref{upper bound delay gradients}. 
\begin{itemize}
\item At first, the negative influence of data heterogeneity is decoupled from the delay, and the negative impact of data heterogeneity is decreased to the same level as the SFL case \eqref{synchronous updates}. This improvement is primarily attributed to reusing delayed gradients \eqref{AFL aggregation rules 2}, which ensures each client participates equally.

\item  Furthermore, the strategy of reusing delayed gradients eliminates the correlation among clients on the training performance. It is evident that smaller $E[\tau_i(t)]$ from any client leads to a smaller upper bound, namely better training performance.

\end{itemize}

\subsection{The Performance Comparison Between PSURDG and AUDG}
Though the proposed aggregation rule of reusing delayed gradients in \eqref{AFL aggregation rules 2} shows excellent convergence characteristics discussed above, it comes with the cost of amplifying the negative impact caused by delays, as evidenced in the last term of two upper bounds in \eqref{upper bound delay gradients} and \eqref{upper bound reusing gradients}. Hence, a performance comparison between AUDG and PSURDG is needed, and their performance gap, denoted as $\Theta$, is defined as the difference between the left-hand side of upper bounds \eqref{upper bound reusing gradients} and \eqref{upper bound delay gradients}, shown as 
\begin{align}
&\Theta = PSURDG(upperbound) - AUDG(upperbound) = \notag \\
&(N-E[|\mathcal{I}_t|]) [\frac{ \eta^2 G^2 L}{2} \sum_{i=1}^N \lambda_i E[\tau_i(t)] - (\frac{3}{2}LR^2 + \frac{2L-\mu}{2}\phi^2)] \notag
\end{align}
Given that $N-E[|\mathcal{I}_t|]$ is non-negative, the sign of $\Theta$ is determined by $\frac{ \eta^2 G^2 L}{2} \sum_{i=1}^N \lambda_i E[\tau_i(t)] - (\frac{3}{2}LR^2 + \frac{2L-\mu}{2}\phi^2)$. Regarding the same delay situation, namely the fixed $\sum_{i=1}^N \lambda_i E[\tau_i(t)]$, PSURDG is superior when the degree of data heterogeneity is large, which is consistent with our initial motivations of reusing the delayed gradients. Besides, the strategy of reusing the gradients is better in training performance when $\frac{ \eta^2 G^2 L}{2} \sum_{i=1}^N \lambda_i E[\tau_i(t)] \leq \frac{3}{2}L R^2 + \frac{L-\mu}{2} \phi^2$, corresponding to the scenario with relatively small delays among clients. When $\Theta$ is less than $0$, the theoretical upper bound of PSURDG is lower than that of AUDG, indicating that PSURDG outperforms AUDG. Based on the above analysis, the delayed information is not always useless, and it surprisingly enhances the training performance for some specific cases, which is opposite to the prevailing belief that outdated data is invariably harmful.

\section{Simulation Results}
In the simulation section, we validate the efficacy of two discussed AFL schemes, AUDG and PSURDG, through image classification tasks. In our simulation configuration, four clients participate in the learning process. The target is to train the model to proficiently identify all ten possible patterns, namely ten labels in the MNIST dataset.

At first, we simulate the SFL with Non-IID data sets, where there is no delays for each client. To satisfy the convexity assumption, an over-parameterized Convolution Neural Network (Over-CNN) is employed, and it has two convolution layers and two fully connected layers with totally 663160 parameters. Additionally, to underscore the impact of the number of parameters, we conduct a comparative simulation with a standard CNN comprising 21,840 parameters. This alternate CNN configuration maintains equivalence with the over-parameterized model, encompassing two convolution layers and two fully connected layers. Moreover, we conduct simulations for both IID and Non-IID scenarios. In the IID setting, each client is equipped with an identical set of 25,000 data samples, ensuring uniformity in their local optimal parameters. In contrast, in the Non-IID setting, each client possesses a distinct set of 6,250 data samples, resulting in diverse data distributions across clients. For both scenarios, a fixed test dataset consisting of 10,000 samples is employed, and the training process comprises 50 iterations, with one gradient descent step executed during each local update. 

\begin{figure}[!t]
\centering
\includegraphics[width=2.4in, trim = 30 5 20 20] {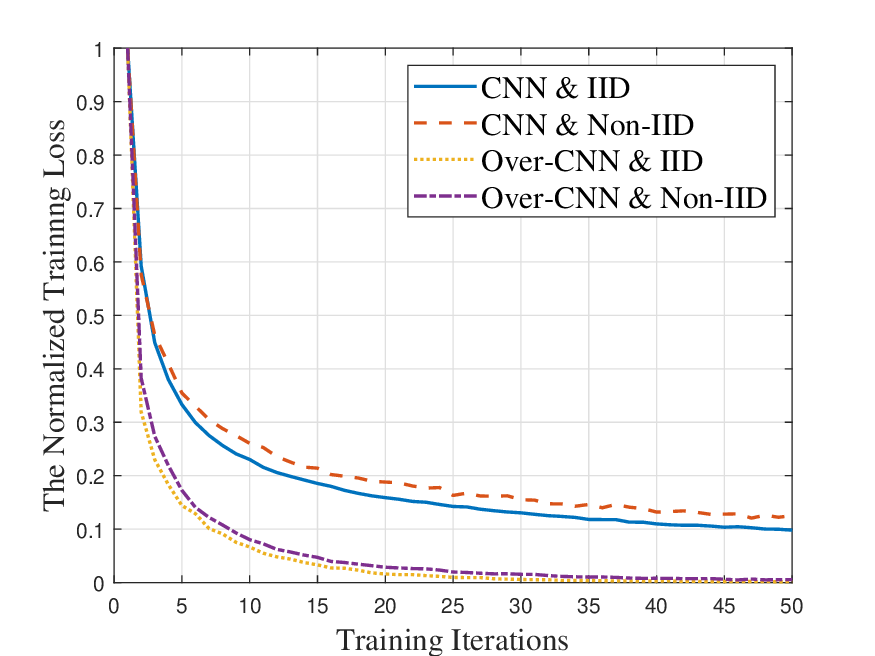}
\captionsetup{font={small}}
\caption{The Evolution of the Normalized Loss Values in SFL}
\label{SFLLoss}
\end{figure}

The evolution of the normalized training loss during the training process is shown in Fig.\ref{SFLLoss}, and Table.\ref{Performance Comparison in SFL} gives the final attained accuracy and loss values of four cases ,where the normalized loss is the loss divided by their corresponding maximum ones. In  Fig.\ref{SFLLoss}, the loss for all cases consistently decrease with increasing iterations, signifying the convergence of the parameters undergoing training. Generally, the over-parameterized CNN shows superior training performance than normal CNN due to more parameters. Additionally, the Non-IID local data sets always exhibits lower accuracy and larger loss values, especially in the normal CNN. Nevertheless, the performance degradation due to data heterogeneity is very limited in the over-parameterized CNN, since the normal CNN needs more iterations to achieve convergence. This observation is consistent with our conclusion from \eqref{synchronous updates},  underscoring that Non-IID datasets in SFL primarily impede training speed without fundamentally obstructing the attainment of theoretical optimal parameters. 

\begin{table}
\centering
\captionsetup{font={small}}
\caption{Performance Comparison in SFL}
\label{Performance Comparison in SFL}
\begin{tabular}{|c|c|c|}
\hline
Setting & Accuracy & The Normalized Loss Values \\ \hline
Over-CNN  \& IID    & 94.01\% & 0.0103\\ \hline
CNN \& IID    & 87.15\% & 0.0966 \\ \hline
Over-CNN  \&  Non-IID    & 93.90\% & 0.0106\\ \hline
CNN  \&  Non-IID    & 85.01\% & 0.1222 \\ \hline
\end{tabular}
\end{table}

To simulate the asynchronous case, according to \cite{gu2021modeling}, we assume each client have the different probabilities of successfully sending parameters back to the center in each iteration, denoted as $\varphi_i$, following an independent and identically distributed Bernoulli process. In accordance with the above settings, the average delay of $client_i$ is $\frac{1}{\varphi_i} -1 $. To explore the impact of the delay introduced by an individual client on the overall training performance, we change the average delay of $client_1$ to observe the change of final accuracy and training loss, while remaining the average delay of other clients as 1 all the time, namely $\varphi_2 = \varphi_2 = \varphi_4 = 0.5$. Due to the randomness of successful transmission, all the results presented below are the average of 10 times Monte Carlo simulations. 

\begin{figure}[!t]
\centering
\includegraphics[width=2.4in, trim = 30 5 20 20] {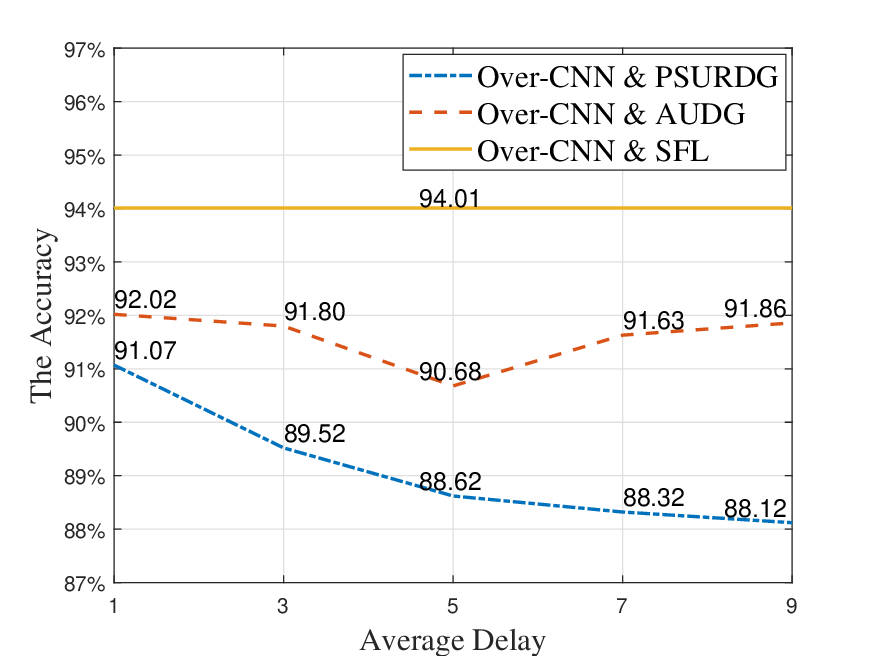}
\captionsetup{font={small}}
\caption{Accuracy: Over-parameterized CNN \& IID }
\label{Accuracy Over IID}
\end{figure}

\begin{figure}[!t]
\centering
\includegraphics[width=2.4in, trim = 30 5 20 20] {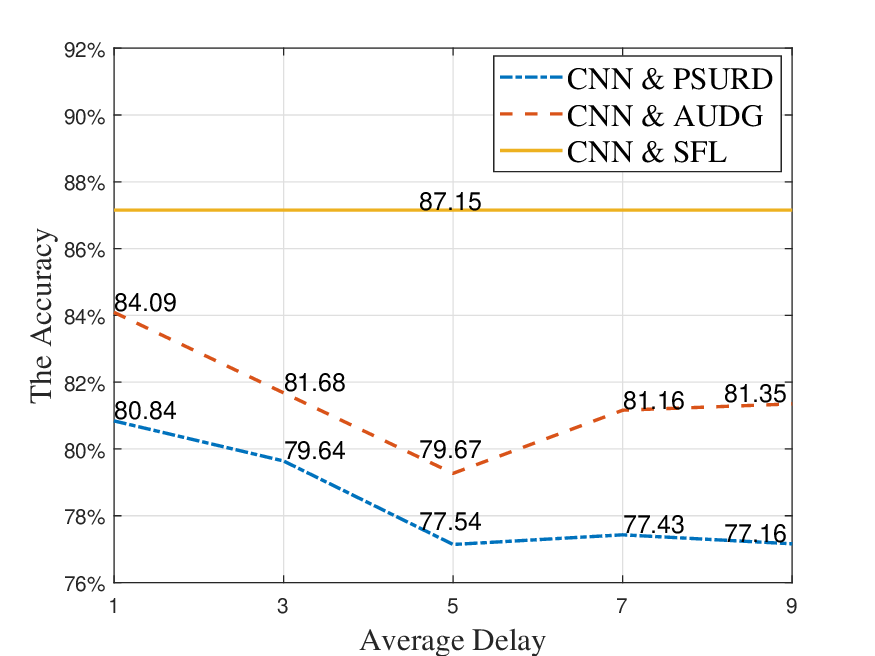}
\captionsetup{font={small}}
\caption{Accuracy: Normal CNN \& IID }
\label{Accuracy IID}
\end{figure}

Initially, we concentrate on scenarios only involving IID local datasets, corresponding to $\phi = 0$ in our theoretical analysis. The simulation settings are identical to those of the IID case in the SFL case, where each client is endowed with the same 25,000 data samples. Both the over-parameterized and the normal CNN are simulated and compared. As to the over-parameterized CNN, Fig.\ref{Accuracy Over IID} and Table.\ref{Accuracy Difference IID IID} exhibits the variation of final attained accuracy after 50 iterations with different average delays of $client_1$, and the corresponding training loss is represented in Table.\ref{Loss over CNN IID}. Fig.\ref{Accuracy IID} and Table.\ref{Loss CNN IID} shows the results of the normal CNN. Generally, the results of training loss from Table.\ref{Loss over CNN IID} and Table.\ref{Loss CNN IID} is consistent with the curves in Fig.\ref{Accuracy Over IID} and Fig.\ref{Accuracy IID}, respectively. With varying average delays, the over-parameterized CNN consistently outperforms the normal CNN for both aggregation rules. Notably, in AUDG, accuracy from the over-parameterized CNN decreases with increasing average delay until the average delay reaches 5, after which it starts to rise. This observation aligns with our theoretical analysis of \eqref{upper bound delay gradients}, and we can observe that the successful transmission of $client_1$ with excessively high latency lower accuracy and increase loss during the training. This alignment between theoretical expectations and experimental findings reinforces the validity of our analytical model. Therefore, further increasing the average delay will decrease the participation times of $client_1$, consequently leading to better training performance. Remarkably, this trend is also observable in the normal CNN, though the convexity assumption is not strictly satisfied. 

As to PSURDG, its accuracy exhibits a monotonically decreasing trend as the average delay increases, consistent with our analytical findings in \eqref{upper bound reusing gradients}. With IID local datasets, the strategy of not reusing delayed gradients is superior in terms of accuracy and training loss, which aligns with our conclusion that reusing the delayed gradient is inefficient when the data heterogeneity is small. According to the theoretical performance gap indicator $\Theta$, the performance gap should increase with the average delay of $client_1$ increasing. However, this theoretical expectation is only validated in the over-parameterized CNN, while the normal CNN diverges from this trend. This discrepancy can be attributed to the non-strict satisfaction of the convexity assumption, arising from the differences in the number of parameters between the two models.

\begin{table}
\centering
\captionsetup{font={small}}
\caption{The Accuracy Difference Between PSURDG and AUDG \& IID}
\label{Accuracy Difference IID IID}
\begin{tabular}{|c|c|c|c|c|c|}
\hline
Average Delay & 1 & 3 & 5 & 7 & 9 \\ \hline
Over-CNN  &-0.95\% 	&-2.28\%  &-2.06\% 	&-3.31\% 	&-3.73\% 
 \\ \hline
CNN   &-3.25\% 	&-2.19\%  &-1.53\% 	&-3.73\% 	&-4.18\% \\ \hline
\end{tabular}
\end{table}

\begin{table}
\centering
\captionsetup{font={small}}
\caption{The Final Training Loss of Over-parameterized CNN \& IID}
\label{Loss over CNN IID}
\begin{tabular}{|c|c|c|c|c|c|}
\hline
Average Delay & 1 & 3 & 5 & 7 & 9 \\ \hline
PSURDG  & 0.1588 & 0.1732 & 0.1949 & 0.1997 & 0.2292 \\ \hline
AUDG   & 0.1411 & 0.1543 & 0.1691 & 0.1593 & 0.1568 \\ \hline
Difference     & 0.0177 & 0.0188 & 0.0258 & 0.0404 & 0.0724 \\ \hline
\end{tabular}
\end{table}

\begin{table}
\centering
\captionsetup{font={small}}
\caption{The Final Training Loss of Normal CNN \& IID}
\label{Loss CNN IID}
\begin{tabular}{|c|c|c|c|c|c|}
\hline
Average Delay & 1 & 3 & 5 & 7 & 9 \\ \hline
PSURDG  & 0.3736 & 0.3751 & 0.4079 & 0.4164 & 0.4216 \\ \hline
AUDG  & 0.3361 & 0.3407 & 0.3926 & 0.3741 & 0.3729 \\ \hline
Difference     & 0.0385 & 0.0343 & 0.0153 & 0.0423 & 0.0487 \\ \hline
\end{tabular}
\end{table}

\begin{table}
\centering
\captionsetup{font={small}}
\caption{The Number of Samples with Three Non-IID Settings}
\label{Number of Data Samples}
\begin{tabular}{|c|c|c|c|c|c|}
\hline
 & $client_1$ & $client_2$ & $client_3$ & $client_4$ \\ \hline
Small Non-IID  &6250	&6250  &6250 	&6250  \\ \hline
Medium Non-IID  &10000 	&5000  &5000 	&5000 \\ \hline
Large Non-IID   &17500 	&2500  &2500 	&2500 \\ \hline
\end{tabular}
\end{table}

Subsequently, we focus on the Non-IID local data sets. Given that the over-parameterized CNN has been established as superior to the normal CNN, our simulation focuses on the over-parameterized CNN within the Non-IID scenario. Additionally, three distinct settings have been prepared to investigate the adverse impact of data heterogeneity on training performance, denoted as Small, Medium, and Large Non-IID, representing different degrees of data heterogeneity shown in Table.\ref{Number of Data Samples}. The test dataset and other parameters remain consistent with those employed in the IID simulations. Generally, the result of training loss aligns consistently with the accuracy trend in corresponding figures for each Non-IID data setting. In contrast to the results obtained from the IID experiments, the training performance of both learning schemes shows inferior performance, manifested through lower accuracy and higher training loss in corresponding average delay, and this observation can be attributed to the adverse impact introduced by data heterogeneity. Correspondingly, the more significant degree of data heterogeneity demonstrates an inferior training performance, reflected through a meticulous comparison of accuracy and training loss at each average delay for PSURDG and AUDG. Diverging from the results of IID local datasets, the accuracy under the both aggregation rules consistently exhibits a monotonically decreasing trend with escalating average delays across three distinct Non-IID settings, which can be attributed to the deleterious impact introduced by the inherent Non-IID characteristics of the data. 

\begin{figure}[!t]
\centering
\includegraphics[width=2.4in, trim = 30 5 20 20] {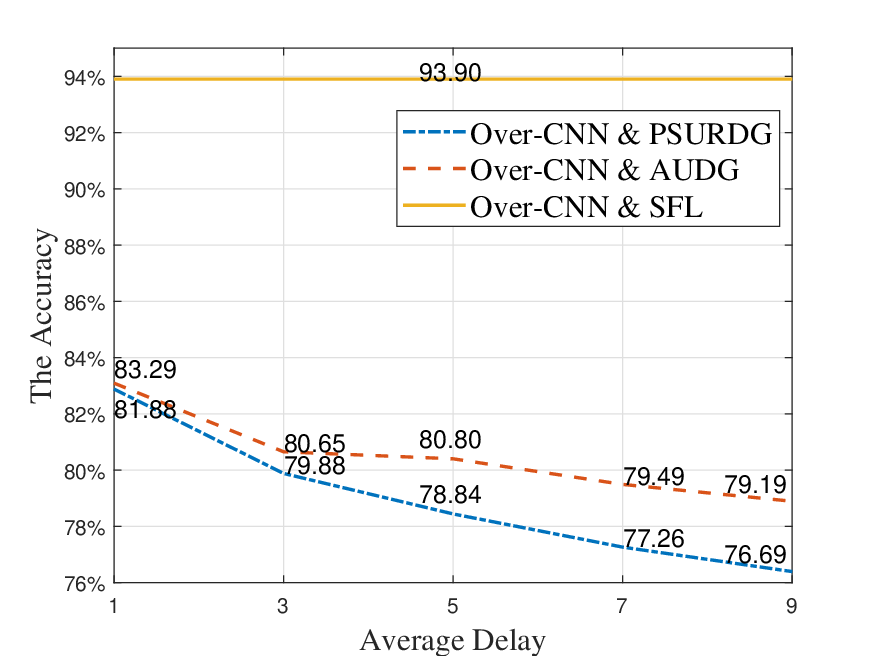}
\captionsetup{font={small}}
\caption{Accuracy: Over-parameterized CNN \& Small Non-IID }
\label{Accuracy Over Non-IID Small}
\end{figure}

\begin{table}
\centering
\captionsetup{font={small}}
\caption{The Final Training Loss: Small Non-IID Data Sets}
\label{Loss Small Non-IID}
\begin{tabular}{|c|c|c|c|c|c|}
\hline
Average Delay & 1 & 3 & 5 & 7 & 9 \\ \hline
PSURDG & 0.5238 &0.5650  &0.6027 &0.6217 &0.6429 \\ \hline
AUDG  & 0.5033 &0.5413 &0.5515 &0.5627 &0.5738 \\ \hline
Difference     & 0.0205 &0.0237 &0.0512 &0.0590 &0.0691 \\ \hline
\end{tabular}
\end{table}

\begin{figure}[!t]
\centering
\includegraphics[width=2.4in, trim = 30 5 20 20] {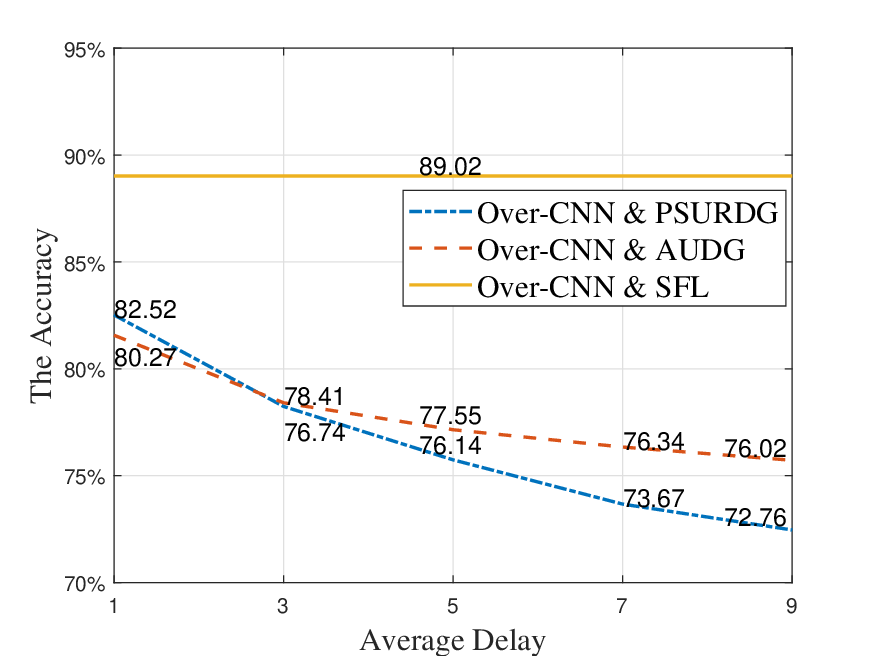}
\captionsetup{font={small}}
\caption{Accuracy: Over-parameterized CNN \& Medium Non-IID }
\label{Accuracy Over Non-IID Medium}
\end{figure}

\begin{table}[]
\centering
\caption{The Final Training Loss: Medium Non-IID Data Sets}
\label{Loss Medium Non-IID}
\begin{tabular}{|c|c|c|c|c|c|}
\hline
Average Delay & 1 & 3 & 5 & 7 & 9 \\ \hline
PSURDG & \textbf{0.4696} &0.5918 &0.6631 &0.7246 &0.7740 \\ \hline
AUDG   & \textbf{0.4803} &0.5779 &0.5995 &0.6459 &0.6606 \\ \hline
Difference     &\textbf{-0.0106} &0.0139 &0.0636 &0.0788 &0.1134 \\ \hline
\end{tabular}
\captionsetup{font={small}}
\end{table}

\begin{figure}[!t]
\centering
\includegraphics[width=2.4in, trim = 30 5 20 20] {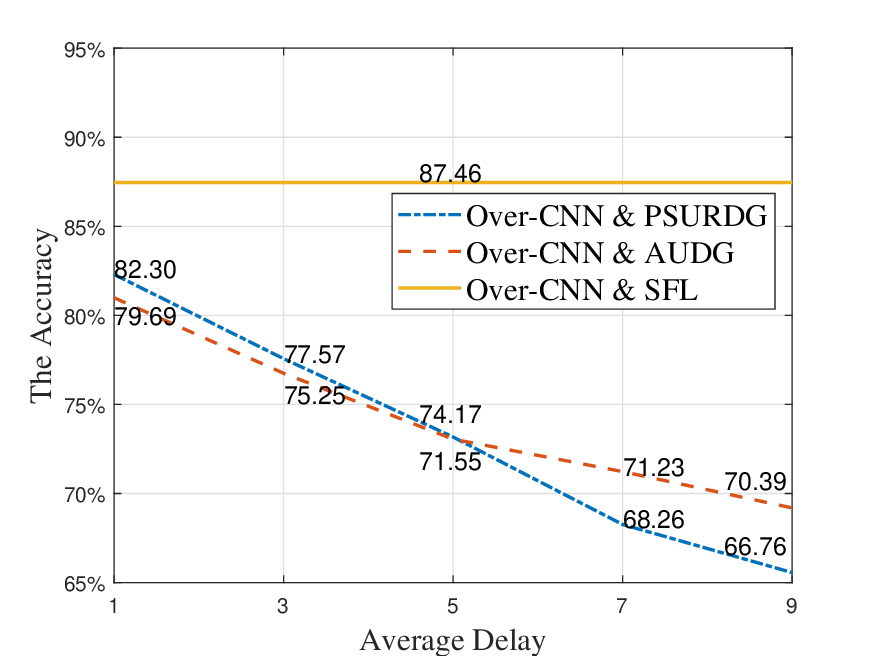}
\captionsetup{font={small}}
\caption{Accuracy: Over-parameterized CNN \& Large Non-IID }
\label{Accuracy Over Non-IID Large}
\end{figure}

\begin{table}
\centering
\captionsetup{font={small}}
\caption{The Final Training Loss: Large Non-IID Data Sets}
\label{Loss Large Non-IID}
\begin{tabular}{|c|c|c|c|c|c|}
\hline
Average Delay & 1 & 3 & 5 & 7 & 9 \\ \hline
PSURDG & \textbf{0.4781} &\textbf{0.6601} 	&\textbf{0.7287} &0.8219 &0.8418 \\ \hline
AUDG  & \textbf{0.4913} &\textbf{0.6699} &\textbf{0.7314} 	&0.7699 	&0.7745 \\ \hline
Difference     & \textbf{-0.0131} 	&\textbf{-0.0098} 	&\textbf{-0.0026} 	&0.0520 	&0.0672 \\ \hline
\end{tabular}
\end{table}

In detail, Figure \ref{Accuracy Over Non-IID Small} delineates the evolution of accuracy related to the escalating average delay for two AFL schemes and the SFL case serving as the benchmark. Table.\ref{Loss Small Non-IID} shows the corresponding result under Small Non-IID setting, and the strategy of reusing the delayed gradient is shown to be inefficient in this case. Figure \ref{Accuracy Over Non-IID Medium} and Table.\ref{Loss Medium Non-IID} present the training outcomes under the Medium data heterogeneity, and PSURDG has higher accuracy and lower training loss when the average delay is 1 for this case. Additionally, the superiority of the strategy of reusing the delayed gradient is also can be observed when the average delay at 1, 3, and 5 for Large Non-IID case, evidenced in Figure \ref{Accuracy Over Non-IID Large} and Table.\ref{Loss Large Non-IID}. These observations are consistent with our theoretical conclusion that reusing the delayed gradient is efficient in the scenario with large data heterogeneity and smaller average delays. Finally, Table.\ref{Accuracy Difference Non-IID IID} presents the accuracy difference between PSURDG and AUDG at different average delays and data heterogeneity, and we can find that the accuracy difference exhibits a monotonically increasing trend with the rise in average delay and an ascent with the augmentation of data heterogeneity. In other words, reusing the delayed gradient is improper for scenarios characterized by larger average delays.

\begin{table}
\centering
\captionsetup{font={small}}
\caption{The Accuracy Difference Between PSURDG and AUDG \& Non-IID}
\label{Accuracy Difference Non-IID IID}
\begin{tabular}{|c|c|c|c|c|c|}
\hline
Average Delay & 1 & 3 & 5 & 7 & 9 \\ \hline
Small Non-IID  &-0.21\% 	&-0.77\%  &-1.96\% 	&-2.23\% 	&-2.51\% \\ \hline
Medium Non-IID  &\textbf{0.95}\% 	&-0.18\%  &-1.42\% 	&-2.67\% 	&-3.26\% \\ \hline
Large Non-IID   &\textbf{1.31}\% 	&\textbf{0.82}\%  &\textbf{0.12}\% 	&-2.97\% 	&-3.63\% \\ \hline
\end{tabular}
\end{table}

\section{Conclusion}
In this paper, we theoretically analyzed the impact of both data heterogeneity and delays in training performance when applying Asynchronous Federated Learning in edge intelligence, where delayed information invariably participates in the aggregation process. In the SFL scenarios, the negative influence only introduced by data heterogeneity is investigated, which is proved to need more iterations to obtain the optimal parameters. In the asynchronous case, the learning scheme AUDG with the aggregation rule of only using the received delayed gradients to update is proposed and analyzed at first. In AUDG, the negative influence of data heterogeneity is proved to be correlated with delays, and all clients are mutually influenced. Besides, a new learning scheme, PSURDG, with the aggregation rule of reusing the delayed gradients, is also investigated, which is proved to be superior to the AUDG in scenarios with smaller delays and larger data heterogeneity. The efficiency of PSURDG also breaks the previous recognition that delayed information has always been useless.  In future works, a new delay adaptive Asynchronous Federated Learning scheme is considered to be constructed in the scenario where the communication channel is limited and the storage space is abundant.
\bibliographystyle{IEEEtran}
\bibliography{IEEEabrv,Ref.bib}

\end{document}